\documentclass[12pt]{article}
\usepackage{amsmath,graphicx}
\usepackage{cite}
\usepackage{amsmath,amssymb,amsfonts}
\usepackage{algorithmic}
\usepackage{graphicx}
\usepackage{textcomp}
\usepackage{hyperref}
\usepackage{enumitem}
\setlist{nosep, leftmargin=14pt}
\usepackage{siunitx}
\usepackage{mwe} 
\usepackage{subcaption}
\usepackage{authblk}

\usepackage{enumitem}
\setlist{nosep, leftmargin=14pt}

\title{Uncertainty-guided annotation enhances segmentation with the human-in-the-loop}
\author[1]{Nadieh Khalili\thanks{\textsuperscript{$\star$} Corresponding author: email@example.com}}
\author[1]{Joey Spronck}
\author[1]{Francesco Ciompi}
\author[1,2]{Jeroen van der Laak}
\author[1]{Geert Litjens}

\affil[1]{Computational Pathology Group, Department of Pathology, Radboud University Medical Center, Nijmegen, The Netherlands}
\affil[2]{Center for Medical Image Science and Visualization, Linköping University, Linköping, Sweden}

\begin{document}
\maketitle
\begin{abstract}

Deep learning algorithms, often critiqued for their 'black box' nature, traditionally fall short in providing the necessary transparency for trusted clinical use. This challenge is particularly evident when such models are deployed in local hospitals, encountering out-of-domain distributions due to varying imaging techniques and patient-specific pathologies. Yet, this limitation offers a unique avenue for continual learning. The Uncertainty-Guided Annotation (UGA) framework introduces a human-in-the-loop approach, enabling AI to convey its uncertainties to clinicians, effectively acting as an automated quality control mechanism. UGA eases this interaction by quantifying uncertainty at the pixel level, thereby revealing the model's limitations and opening the door for clinician-guided corrections. We evaluated UGA on the Camelyon dataset for lymph node metastasis segmentation which revealed that UGA improved the Dice coefficient (DC), from 0.66 to 0.76 by adding 5 patches, and further to 0.84 with 10 patches. To foster broader application and community contribution, we have made our code accessible at \hyperlink{https://github.com/nadieh/Collaborative_segmentation}{GitHub}.

\end{abstract}

\section{Introduction}
Domain shift is a critical issue in medical imaging, exacerbated by factors like vendor inconsistencies, patient cohorts, different pathologies, and variable workflows. This challenge is even more prominent in digital pathology due to lab procedure variability, including tissue staining procedures. Previously several augmentation techniques addressed staining variability by synthesizing different color distributions during training \cite{wagner2021structure,tellez2019quantifying,faryna2021tailoring}. 
Even with the notable gains achieved through augmentation, full generalizability still eludes many deep neural networks, largely due to unpredictable scenarios. In unpredictable situations, deep learning models may yield incorrect predictions without alerting the user. Therefore, it is imperative to apply deep learning techniques that communicate uncertainty to clinicians and make them part of a feedback loop. 
\begin{figure}[htbp!]
    \centering
        \includegraphics[ width = 0.9\linewidth]{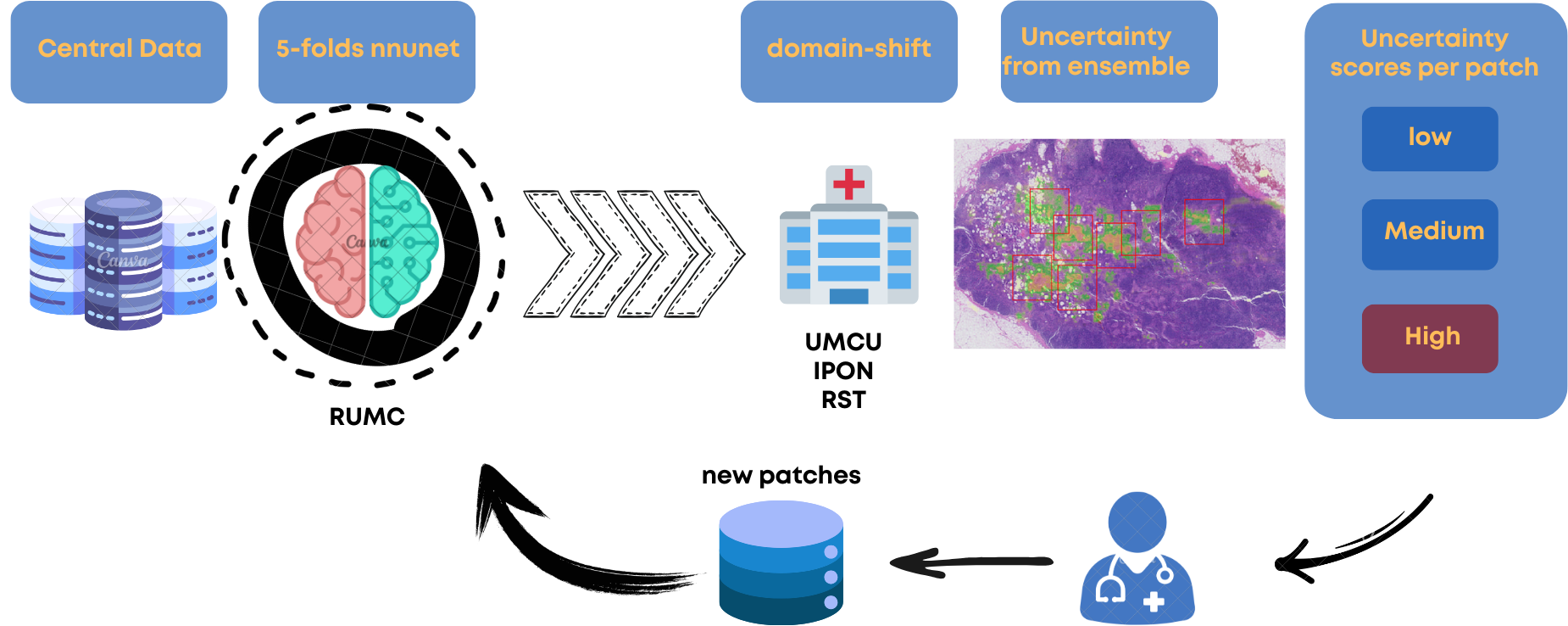}
    \caption{The collaborative segmentation method using UGA is illustrated. The segmentation model trained on the central dataset (RUMC) is applied to five different centers including RUMC, UMCU, IPON, CWZ and RST. The network quantifies the uncertainty per patch using ensembles of nnU-Net and is sorted in descending order from the most uncertain to the least uncertain patches. The human-in-the-loop process reviews the cases with the highest uncertainty (only 5 patches) and corrects segmentation. The new version of the model is trained on a combination of central and local data.}
    \label{fig:method}
\end{figure}
Several deep learning techniques have demonstrated promising results in accurately gauging general and pixel-level uncertainties and detecting out-of-domain samples \cite{dolezal2022uncertainty, lakshminarayanan2017simple}. Deep ensembles, as highlighted in Lakshminarayanan et al. offer enhanced robustness in uncertainty quantification, specifically in digital pathology \cite{linmans2023predictive}. These models, built as ensembles of several deep learning model instances, leverage the variance among these instances as a measure of uncertainty. Such ensembles of instance segmentation using quantifies segmentation uncertainty at tissue level. nnU-Net, is currently the leading algorithm for radiological image segmentation, which has been adeptly modified for use in pathology \cite{isensee2021nnu,spronck2023nnunet}.

In this study, we present a human-in-the-loop framework that utilizes model uncertainty as a critical tool for enhancing segmentation quality in pathology. Our framework, Uncertainty-Guided Annotation (UGA), integrates clinician expertise directly into the learning process enabling continuous model refinement and learning to improve generalization. By selecting the areas of high uncertainty, UGA involves clinicians to contribute targeted corrections, thereby refining the model's understanding in segmentation tasks. This innovative approach redefines the training process, as the model evolves through a combination of original data and clinician-provided corrections. We validate this method on the Camelyon 16 and 17 datasets, specifically focusing on the challenging task of metastasis segmentation in whole-slide images.

\section{Data}
\begin{figure}[htbp!]
  \centering\includegraphics[scale = 0.5]{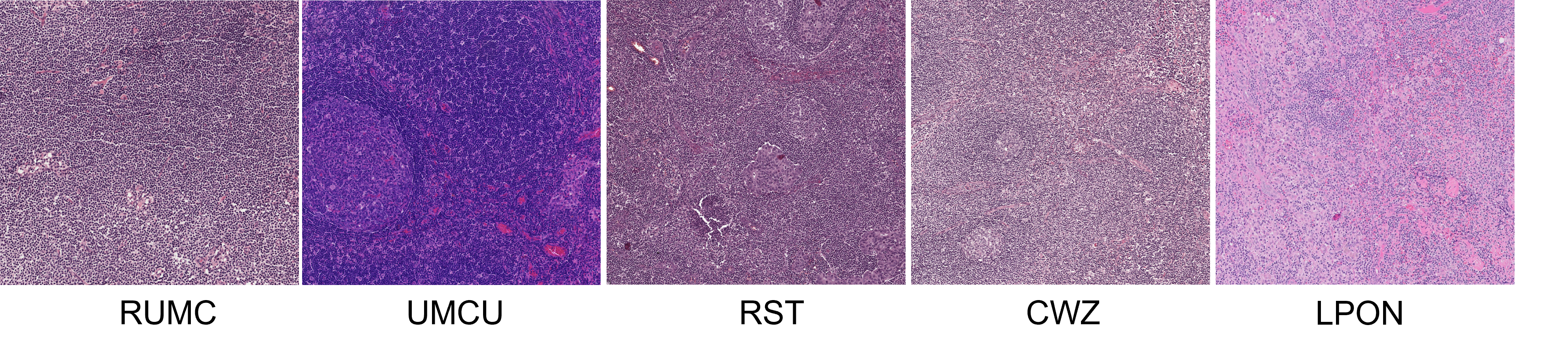}

    \caption{Overview of the color variation observed among different centers in the RUMC, UMCU, RST, CWZ, and LabPON datasets.
    }
    \label{fig:cam17}
\end{figure}

For the training and evaluation of our model, we utilized the publicly available datasets Camelyon16 and Camelyon17. Camelyon16 features whole slide images (WSIs) of lymph nodes from two medical centers: Radboud University Medical Center (RUMC) and University Medical Center Utrecht (UMCU). We used 137 WSIs only from RUMC for the training and validation of our baseline model. Camelyon17 offers a more diverse dataset comprising WSIs from five different medical centers: RUMC, UMCU, Rijnstate Hospital (RST), Canisius-Wilhelmina Hospital (CWZ), and Laboratorium Pathologie Oost-Nederland (LabPON). In addition to the 50 WSIs with pixel-level lesion annotations that Camelyon17 provides, we supplemented it with 50 benign slides that effectively serve as manual annotations at no additional cost due to their lesion-free status. Notably, the dataset visualizes the variations in staining techniques across these five centers see Figure \ref{fig:cam17}. Of this set, 75 WSIs were used for applying uncertainty and sampling patches with the highest uncertainty, while 25 WSIs remained untouched for testing. Details about data can be found in the Camelyon challenge website \hyperlink{https://camelyon16.grand-challenge.org/Data/}{https://camelyon16.grand-challenge.org/Data/} and \hyperlink{https://camelyon17.grand-challenge.org/}{https://camelyon17.grand-challenge.org/}.

\section{Uncertainty-guided annotation sampling}
The proposed human-in-the-loop model quantifies the uncertainty using ensembles of pathology-adapted nnU-Net\cite{spronck2023nnunet}. We used this modified version of nnU-Net, given that the original nnU-Net was initially designed for radiology applications and faced limitations with Whole Slide Images (WSIs). Due to WSIs' large file sizes, we leveraged the WholeSlideData Python library\footnote{https://github.com/DIAGNijmegen/pathology-whole-slide-data} for efficient extraction of $1024\times 1024$ patches and their associated annotations from the training and validation sets, at a spatial resolution of \SI{1}{\micro\metre/px}. To reduce the imbalance, inherently present in the WSIs representation of metastatic and benign regions, a 1:4 sampling ratio was employed. During the training phase, we set the batch size to 2 and applied z-score normalization to mitigate color discrepancies, as suggested by Spronck et al. \cite{spronck2023nnunet}. The training was conducted via a 5-fold cross-validation scheme, resulting in five distinct models.

The model was trained on 137 WSIs from RUMC, a subset of the Camelyon16 dataset, and applied to 75 WSIs from the Camelyon17. The uncertainty in our model is measured through the degree of inconsistency among the five cross-validated folds. Specifically, greater variance among the folds correlates with increased per-pixel uncertainty. For each pixel, internal disagreement is calculated using soft cross-entropy loss by contrasting the individual fold's log(softmax) against the collective mean of log(softmax) across the five folds. This procedure effectively establishes a co-adaptive feedback loop between the model and the user, targeted at enhancing overall accuracy. The final evaluation was conducted using Camelyon17's test set. The uncertainty ascertained by summing the individual pixel-level uncertainty scores in $1024\times 1024$ patches. These aggregate scores were compiled for all 75 WSIs, ranked based on each center, and their coordinates were stored for future patch extraction. To enhance uncertainty quality, patches dominated by black or white pixels over 70\% were excluded from the analysis. The patches with the highest average uncertainty were sampled and added to the training dataset. The 25 remaining WSIs from Camelyon17 stayed untouched to serve as an independent test set. It's important to clarify that in this study we simulated the corrections of clinicians through existing annotations (see Fig. \ref{fig:method}).

\section{Experiments and Results}
To evaluate our UGA approach, we initially trained a model exclusively on RUMC data and then assessed its performance on the Camelyon 17 test set. A series of experiments were conducted to compare the effectiveness of UGA with a random sampling strategy. For both strategies, five patches of $1024\times 1024$ pixels were selected from each of the five centers, yielding a total of 25 samples. A criterion was implemented to exclude patches where over $70\%$ is background (white or black). The primary distinction between the two methods is the selection criterion: UGA strategically picks 5, 10 and 20 patches with the highest uncertainty from each center, whereas random sampling indiscriminately selects patches from the foreground areas. Additionally, we augmented each sample 100 times using standard flipping and hue augmentation. Subsequently, we continued the model's training, utilizing both new and old data, with baseline weights serving as the weight initialization. 

\begin{figure}[htp!]
    \centering
        \includegraphics[scale = 1]{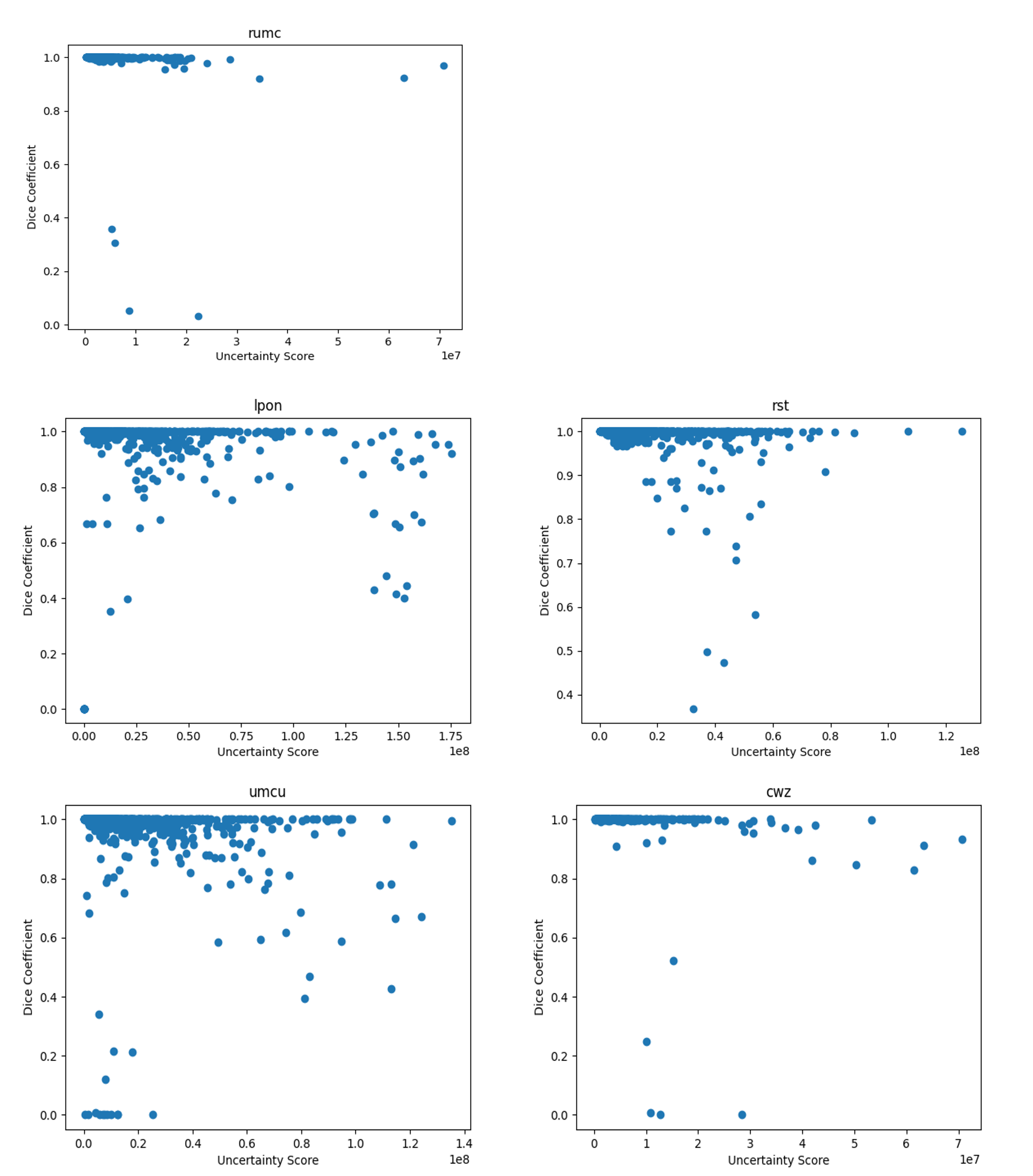}

    \caption{The model trained solely on RUMC data, is applied to datasets from five different centers. The graph showcases both aggregated patch-level uncertainty values and corresponding DC.}
    \label{fig:uncertaint}
\end{figure}

\begin{figure}[htp]
  \begin{subfigure}{0.3\textwidth}
    \includegraphics[scale=0.3]{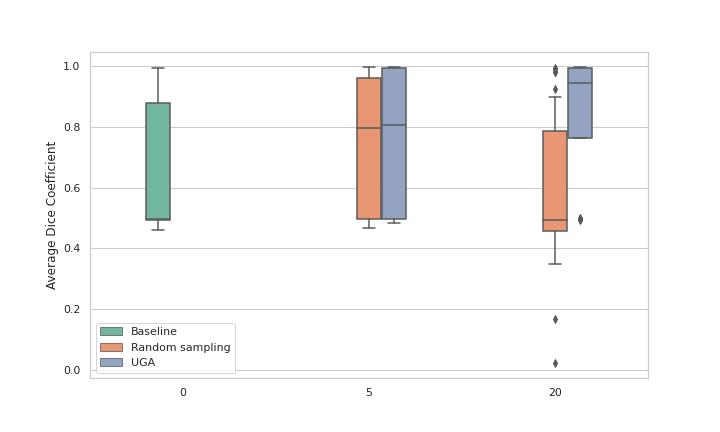}

    \caption{}
\label{fig:patch}

  \end{subfigure}
  \begin{subfigure}{0.45\textwidth}
    \includegraphics[scale=0.3]{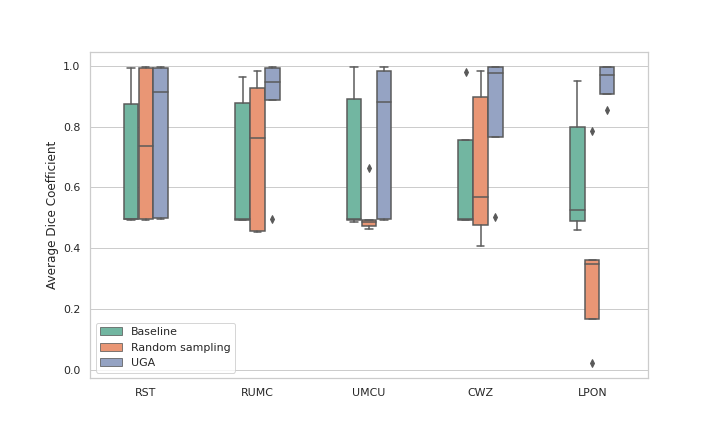}
    \caption{}
\label{fig:centers}

  \end{subfigure}
  \begin{subfigure}{0.3\textwidth}
    \includegraphics[scale=0.3]{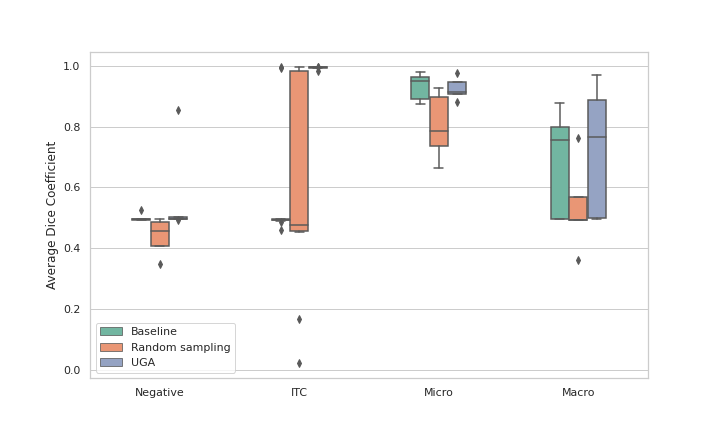}
    \caption{}
    \label{fig:cancer}

  \end{subfigure}
  \caption{The performance of the segmentation models was quantified using DC and compared across three training strategies: the UGA approach, random sampling, and a baseline model trained exclusively on data from RUMC. (a) Segmentation performanc of the baseline model, continue training on additional training on 5 and 10 patches. (b) Segmentation performance across RST, RUMC, UMCU, CWZ and Lpon centers. (c) Segmentation performance for different cancer types, including negative samples, isolated tumor cells (ITC), micro-metastases, macro-metastases.}
\label{fig:boxplots}

\end{figure}

The visualization of uncertainty values for patches from each center, alongside their corresponding DC, is presented in Figure \ref{fig:uncertaint}. Notably, the lowest uncertainty levels are observed for patches from RUMC, which aligns with RUMC being the primary training dataset for the network. Figure \ref{fig:boxplots} shows that the baseline model exhibited substantial performance gains with the incremental addition of training patches. Average DC improved from 0.66 to 0.76 and 0.84, adding 5 patches, and further increased after including 20 additional patches using UGA. The models that received additional patches using UGA sampling outperformed random sampling. Figure \ref{fig:centers} illustrates UGA sampling consistently improved segmentation performance in compar to random sampling at each center. The advantage of UGA sampling is further underlined when analyzing different tissue types in Figure \ref{fig:cancer}. The UGA approach was especially adept at identifying ITCs, which are frequently undetected by AI due to their scarcity.

\section{Conclusion}

We have developed a novel human-in-the-loop strategy for WSIs segmentation that harnesses model uncertainty as a diagnostic tool for highlighting problematic, out-of-domain areas. This approach enriches the collaborative interaction between clinicians and AI models by signaling areas of high uncertainty as likely errors in the model's predictions. Clinicians can correct these errors, enabling the model to improve through retraining. The preliminary results affirm that incorporating even a few of these corrected high-uncertainty areas enhances the model's performance.

Generally, for pathologists, segmenting cancer from lymph nodes is considered a simpler task compared to the intricacies of multiple cancer grading tasks. This relative simplicity is reflected in AI models, which demonstrate lower uncertainty in such segmentation, as illustrated in Figure \ref{fig:uncertaint}. Our future research aims to assess the model in more challenging tasks, where a greater degree of variation is inherent to the task itself.

Deep learning models are often perceived as 'black boxes' due to their lack of transparent communication with users. Incorporating uncertainty measures can enhance this communication, fostering trust between the user and the AI model. This approach not only elucidates model confidence but also guides pathologists to areas of potential model improvement. Unlike traditional active learning methods, which primarily focus on optimizing the training dataset \cite{nath2020diminishing}, our human-in-the-loop methodology enables dynamic, continual learning between the AI system and the pathologist. It allows pathologists to provide targeted feedback on specific regions of interest where the model exhibits uncertainty.

Moreover, our approach is particularly compatible with the federated learning approach, which emphasizes privacy-preserving collaborative model development. In such paradigms, domain experts can interact directly with the local instances of the model—providing targeted corrections and insights—while all patient data remains securely within the confines of the originating healthcare institution. When aligned with federated learning, this human-in-the-loop network not only upholds stringent data privacy regulations by design but also prevents data leakage by avoiding the centralization of sensitive information. These privacy-centric characteristics suggest a promising avenue for future research, aiming to scale and integrate the human-in-the-loop methodology across distributed healthcare environments, thereby enhancing model robustness and patient trust.

%
%
%
%

\bibliographystyle{elsarticle-num}
\bibliography{sample}

\begin{thebibliography}{1}
\expandafter\ifx\csname url\endcsname\relax
  \def\url#1{\texttt{#1}}\fi
\expandafter\ifx\csname urlprefix\endcsname\relax\def\urlprefix{URL }\fi
\expandafter\ifx\csname href\endcsname\relax
  \def\href#1#2{#2} \def\path#1{#1}\fi

\bibitem{wagner2021structure}
S.~J. Wagner, N.~Khalili, R.~Sharma, M.~Boxberg, C.~Marr, W.~de~Back, T.~Peng, Structure-preserving multi-domain stain color augmentation using style-transfer with disentangled representations, in: Medical Image Computing and Computer Assisted Intervention--MICCAI 2021: 24th International Conference, Strasbourg, France, September 27--October 1, 2021, Proceedings, Part VIII 24, Springer, 2021, pp. 257--266.

\bibitem{tellez2019quantifying}
D.~Tellez, G.~Litjens, P.~B{\'a}ndi, W.~Bulten, J.-M. Bokhorst, F.~Ciompi, J.~Van Der~Laak, Quantifying the effects of data augmentation and stain color normalization in convolutional neural networks for computational pathology, Medical image analysis 58 (2019) 101544.

\bibitem{faryna2021tailoring}
K.~Faryna, J.~van~der Laak, G.~Litjens, Tailoring automated data augmentation to h\&e-stained histopathology, in: Medical Imaging with Deep Learning, 2021.

\bibitem{dolezal2022uncertainty}
J.~M. Dolezal, A.~Srisuwananukorn, D.~Karpeyev, S.~Ramesh, S.~Kochanny, B.~Cody, A.~S. Mansfield, S.~Rakshit, R.~Bansal, M.~C. Bois, et~al., Uncertainty-informed deep learning models enable high-confidence predictions for digital histopathology, Nature communications 13~(1) (2022) 6572.

\bibitem{lakshminarayanan2017simple}
B.~Lakshminarayanan, A.~Pritzel, C.~Blundell, Simple and scalable predictive uncertainty estimation using deep ensembles, Advances in neural information processing systems 30 (2017).

\bibitem{linmans2023predictive}
J.~Linmans, S.~Elfwing, J.~van~der Laak, G.~Litjens, Predictive uncertainty estimation for out-of-distribution detection in digital pathology, Medical Image Analysis 83 (2023) 102655.

\bibitem{isensee2021nnu}
F.~Isensee, P.~F. Jaeger, S.~A. Kohl, J.~Petersen, K.~H. Maier-Hein, nnu-net: a self-configuring method for deep learning-based biomedical image segmentation, Nature methods 18~(2) (2021) 203--211.

\bibitem{spronck2023nnunet}
J.~Spronck, T.~Gelton, L.~van Eekelen, J.~Bogaerts, L.~Tessier, M.~van Rijthoven, L.~van~der Woude, M.~van~den Heuvel, W.~Theelen, J.~van~der Laak, et~al., nnunet meets pathology: bridging the gap for application to whole-slide images and computational biomarkers, in: Medical Imaging with Deep Learning, 2023.

\bibitem{nath2020diminishing}
V.~Nath, D.~Yang, B.~A. Landman, D.~Xu, H.~R. Roth, Diminishing uncertainty within the training pool: Active learning for medical image segmentation, IEEE Transactions on Medical Imaging 40~(10) (2020) 2534--2547.

\end{thebibliography}
\end{document}